# Application of three graph Laplacian based semi-supervised learning methods to protein function prediction problem

Loc Tran

University of Minnesota

tran0398@umn.edu

*Abstract:*

*Protein function prediction is the important problem in modern biology. In this paper, the un-normalized, symmetric normalized, and random walk graph Laplacian based semi-supervised learning methods will be applied to the integrated network combined from multiple networks to predict the functions of all yeast proteins in these multiple networks. These multiple networks are network created from Pfam domain structure, co-participation in a protein complex, protein-protein interaction network, genetic interaction network, and network created from cell cycle gene expression measurements. Multiple networks are combined with fixed weights instead of using convex optimization to determine the combination weights due to high time complexity of convex optimization method. This simple combination method will not affect the accuracy performance measures of the three semi-supervised learning methods. Experiment results show that the un-normalized and symmetric normalized graph Laplacian based methods perform slightly better than random walk graph Laplacian based method for integrated network. Moreover, the accuracy performance measures of these three semi-supervised learning methods for integrated network are much better than the best accuracy performance measures of these three methods for the individual network.*

## 1. Introduction

Protein function prediction is the important problem in modern biology. Identifying the function of proteins by biological experiments is very expensive and hard. Hence a lot of computational methods have been proposed to infer the functions of the proteins by using various types of information such as gene expression data and protein-protein interaction networks [1].

First, in order to predict protein function, the sequence similarity algorithms [2, 3] can be employed to find the homologies between the already annotated proteins and theun-annotated protein. Then the annotated proteins with similar sequences can be used to assign the function to the un-annotated protein. That's the classical way to predict protein function [4].

Second, to predict protein function, a graph (i.e. kernel) which is the natural model of relationship between proteinscan also be employed. In this model, the nodes represent proteins. The edges represent for the possible interactions between nodes. Then, machine learning methods such as Support Vector Machine [5], Artificial Neural Networks [4], un-normalized graph Laplacian based semi-supervised learning method [6,14], or neighbor counting method [7] can be applied to this graph to infer the functions of un-annotated protein. The neighbor counting method labels the protein with the function that occurs frequently in the protein's adjacent nodes in the protein-protein interaction network. Hence neighbor counting method does not utilize the full topology of the network. However, the Artificial Neural Networks, Support Vector Machine, andun-normalized graph Laplacian based semi-supervised learning method utilize the full topology of

International Journal on Bioinformatics & Biosciences (IJBB) Vol.3, No.2, June 2013

the network.Moreover, the Artificial Neural Networks and Support Vector Machine are supervised learning methods.

While the neighbor counting method, the Artificial Neural Networks, and the un-normalized graph Laplacian based semi-supervised learningmethod are all based on the assumption that the labels of two adjacent proteins in graph are likely to be the same, SVM do not rely on this assumption. Graphs used in neighbor counting method, Artificial Neural Networks, and the un-normalized graph Laplacian based semi-supervised learningmethod are very sparse.However, the graph (i.e. kernel) used in SVM is fully-connected.

Third, the Artificial Neural Networks method is applied to the single protein-protein interaction network only. However, the SVM method and un-normalized graph Laplacian based semi-supervised learning method try to use weighted combination of multiple networks(i.e. kernels) such as gene co-expression networkand protein-protein interaction network to improve the accuracy performance measures. While [5] (SVM method) determines the optimal weighted combination of networks by solving the semi-definite problem, [6,14] (un-normalized graph Laplacian based semi-supervised learning method) uses a dual problem and gradient descent to determine the weighted combination of networks.

In the last decade, the normalized graph Laplacian [8] and random walk graph Laplacian [9] based semi-supervised learning methods have successfully been applied to some specific classification tasks such as digit recognition and text classification. However, to the best of my knowledge, the normalized graph Laplacian and random walk graph Laplacian based semi-supervised learning methods have not yet been applied to protein function prediction problem and hence their overall accuracy performance measure comparisons have not been done. In this paper, we will apply three un-normalized, symmetric normalized, and random walk graph Laplacian based semi-supervised learning methods to the integrated network combined with fixed weights.These five networksused for the combination are available from [6]. The main point of these three methods is to let every node of the graph iteratively propagates its label information to its adjacent nodes and the process is repeated until convergence [8]. Moreover, since [6] has pointed out that the integrated network combined with optimized weights has similar performance to the integrated network combined with equal weights, i.e. without optimization, we will use the integrated network combined with equal weights due to high time-complexity of these optimization methods. This type of combination will be discussed clearly in the next sections.

We will organize the paper as follows: Section 2 will introduce random walk and symmetric normalized graph Laplacian based semi-supervised learning algorithms in detail.Section 3will show how to derive the closed form solutions of normalized and un-normalized graph Laplacian based semi-supervised learning from regularization framework. In section 4, we will apply these three algorithms to the integrated network of five networks available from [6]. These five networks are network created from Pfam domain structure, co-participation in a protein complex, protein-protein interaction network, genetic interaction network, and network created from cell cycle gene expression measurements. Section 5 will conclude this paper and discuss the future directions of researches of this protein function prediction problem utilizing hypergraph Laplacian.

**Claim:** Random walk and symmetric normalized graph Laplacians have been widely used not in classification but also in clustering [8,13]. In this paper, we will focus on the application of these two graph Laplacians to the protein function prediction problem. The accuracy performance measures of these two methods will be compared to the accuracy performance measure of the un-normalized graph Laplacian based semi-We **do not claim** that the accuracy performance measures of these two methods will be better than the accuracy performance measure of the un-normalized graph Laplacian based semi-supervised learning method (i.e. the published method) in this protein function prediction problem. We just do the comparisons.

To the best of my knowledge, no theoretical framework have been given to prove that which graph Laplacian method achieves the best accuracy performance measure in the classification task. In the other words, the accuracy performance measures of these three graph Laplacian based semi-supervised learning methods depend on the datasets we used. However, in [8], the author have pointed out that the accuracy performance measure of the symmetric normalized graph Laplacian based semi-supervised learning method are better than accuracy performance measures of the random walk and un-normalized graph Laplacian based semi-supervised learning methods in digit recognition and text categorization problems. Moreover, its accuracy performance measure is also better than the accuracy performance measure of Support Vector Machine method (i.e. the known best classifier in literature) in two proposed digit recognition and text categorization problems. This fact is worth investigated in protein function prediction problem.

Again, we do not claim that our two proposed random walk and symmetric normalized graph Laplacian based semi-supervised learning methods will perform better than the published method (i.e. the un-normalized graph Laplacian method)in this protein function prediction problem. At least, the accuracy performance measures of two new proposed methods are similar to or are not worse than the accuracy performance measure of the published method (i.e. the un-normalized graph Laplacian method).

## 2. Algorithms

Given $m$ networks in the dataset, the weights for individual networks used to combine to integrated network are $\frac{1}{m}$.

Given a set of proteins $\{x_1, \ldots, x_l, x_{l+1}, \ldots, x_{l+u}\}$ where $n = l + u$ is the total number of proteins in the integrated network, define c bethe total number of functional classes and the matrix $F \in R^{n \cdot c}$ be the estimated label matrix for the set of proteins $\{x_1, \ldots, x_l, x_{l+1}, \ldots, x_{l+u}\}$, where the point $x_i$ is labeled as sign($F_{ij}$) for each functional class j ($1 \leq j \leq c$). Please note that $\{x_1, \ldots, x_l\}$ is the set of all labeled points and $\{x_{l+1}, \ldots, x_{l+u}\}$ is the set of all un-labeled points.

Let $Y \in R^{n \cdot c}$ the initial label matrix for n proteins in the network be defined as follows

$$Y_{ij} = \begin{cases} 1 & if\ x_i\ belongs\ to\ functional\ class\ j\ and\ 1 \leq i \leq l \\ -1 & if\ x_i\ does\ not\ belong\ to\ functional\ class\ j\ and\ 1 \leq i \leq l \\ 0 & if\ l+1 \leq i \leq n \end{cases}$$

Our objective is to predict the labels of the un-labeled points $x_{l+1}, \ldots, x_{l+u}$. We can achieve this objective by letting every node (i.e. proteins) in the network iteratively propagates its label information to its adjacent nodes and this process is repeated until convergence. These three algorithms are based on three assumptions:

- local consistency: nearby proteins are likely to have the same function
- global consistency: proteins on the same structure (cluster or sub-manifolds) are likely to have the same function
- these protein networks contain no self-loops

Let $W^{(k)}$ represents the $k^{th}$ individual network in the dataset.

### Random walk graph Laplacian based semi-supervised learning algorithm

In this section, we slightly change the original random walk graph Laplacian based semi-supervised learning algorithm can be obtained from [9]. The outline of the new version of this algorithm is as follows

1. Form the affinity matrix $W^{(k)}$ (for each k such that $1 \le k \le m$):
$$W_{ij}^{(k)} = \begin{cases} \exp\{-\frac{||x_i - x_j||}{2*\delta^2}\} & if\ i \neq j \\ 0 & if\ i = j \end{cases}$$

2. Construct $S_{rw} = \frac{1}{m}\sum_{k=1}^{m} D^{(k)^{-1}} W^{(k)}$ where $D^{(k)} = diag(d_1^{(k)}, \ldots, d_n^{(k)})$ and $d_i^{(k)} = \sum_j W_{ij}^{(k)}$

3. Iterate until convergence
$F^{(t+1)} = \alpha S_{rw} F^{(t)} + (1-\alpha)Y$, where $\alpha$ is an arbitrary parameter belongs to [0,1]

4. Let $F^*$ be the limit of the sequence $\{F^{(t)}\}$. For each protein functional class j, label each protein $x_i (l+1 \le i \le l+u)$ as $sign(F_{ij}^*)$

Next, we look for the closed-form solution of the random walk graph Laplacian based semi-supervised learning. In the other words, we need to show that

$$F^* = \lim_{t \to \infty} F^{(t)} = (1-\alpha)(I - \alpha S_{rw})^{-1} Y$$

Suppose $F^{(0)} = Y$, then

$$F^{(1)} = \alpha S_{rw} F^{(0)} + (1-\alpha)Y$$
$$= \alpha S_{rw} Y + (1-\alpha)Y$$

$$F^{(2)} = \alpha S_{rw} F^{(1)} + (1-\alpha)Y$$
$$= \alpha S_{rw}(\alpha S_{rw} Y + (1-\alpha)Y) + (1-\alpha)Y$$
$$= \alpha^2 S_{rw}^2 Y + (1-\alpha)\alpha S_{rw} Y + (1-\alpha)Y$$

$$F^{(3)} = \alpha S_{rw} F^{(2)} + (1-\alpha)Y$$
$$= \alpha S_{rw}(\alpha^2 S_{rw}^2 Y + (1-\alpha)\alpha S_{rw} Y + (1-\alpha)Y) + (1-\alpha)Y$$
$$= \alpha^3 S_{rw}^3 Y + (1-\alpha)\alpha^2 S_{rw}^2 Y + (1-\alpha)\alpha S_{rw} Y + (1-\alpha)Y$$

…

Thus, by induction,

$$F^{(t)} = \alpha^t S_{rw}^t Y + (1-\alpha)\sum_{i=0}^{t-1}(\alpha S_{rw})^i Y$$

Since $S_{rw}$ is the stochastic matrix, its eigenvalues are in [-1,1]. Moreover, since $0 < \alpha < 1$, thus

$$\lim_{t \to \infty} \alpha^t S_{rw}^t = 0$$

$$\lim_{t \to \infty} \sum_{i=0}^{t-1}(\alpha S_{rw})^i = (I - \alpha S_{rw})^{-1}$$

Therefore,

$$F^* = \lim_{t \to \infty} F^{(t)} = (1-\alpha)(I - \alpha S_{rw})^{-1} Y$$

Now, from the above formula, we can compute $F$ directly.

The original random walk graph Laplacian based semi-supervised learning algorithm developed by Zhu can be derived from the modified algorithm by setting $\alpha_i = 0$, where $1 \leq i \leq l$ and $\alpha_i = 1$, where $l + 1 \leq i \leq l + u$. In the other words, we can express $F^{(t+1)}$ in matrix form as follows

$$F^{(t+1)} = I_\alpha S_{rw} F^{(t)} + (I - I_\alpha) Y, \text{ where}$$

I is the identity matrix and $I_\alpha = \begin{bmatrix} 0 & \cdots & 0 & & & \\ \vdots & \ddots & & 0 & & \\ 0 & \cdots & 0 & & & \\ & & & 1 & \cdots & 0 \\ & 0 & & & \ddots & \\ & & & 0 & \cdots & 1 \end{bmatrix}$ ($I_\alpha$ is the diagonal matrix)

**Normalized graph Laplacian based semi-supervised learning algorithm**

Next, we will give the brief overview of the original normalized graph Laplacian based semi-supervised learning algorithm can be obtained from [8]. The outline of this algorithm is as follows

1. Form the affinity matrix $W^{(k)}$ (for each $1 \leq k \leq m$):

$$W_{ij}^{(k)} = \begin{cases} \exp\left\{-\frac{||x_i - x_j||}{2 * \delta^2}\right\} & \text{if } i \neq j \\ 0 & \text{if } i = j \end{cases}$$

2. Construct $S_{sym} = \frac{1}{m} \sum_{k=1}^{m} D^{(k)^{-\frac{1}{2}}} W^{(k)} D^{(k)^{-\frac{1}{2}}}$ where $D^{(k)} = \text{diag}(d_1^{(k)}, \ldots, d_n^{(k)})$ and $d_i^{(k)} = \sum_j W_{ij}^{(k)}$

3. Iterate until convergence
   $F^{(t+1)} = \alpha S_{sym} F^{(t)} + (1 - \alpha) Y$, where is an arbitrary parameter belongs to [0,1]

4. Let $F^*$ be the limit of the sequence $\{F^{(t)}\}$. For each protein functional class j, label each protein $x_i (l + 1 \leq i \leq l + u)$ as $\text{sign}(F_{ij}^*)$

Next, we look for the closed-form solution of the normalizedgraph Laplacian based semi-supervised learning. In the other words, we need to show that

$$F^* = \lim_{t \to \infty} F^{(t)} = (1-\alpha)(I - \alpha S_{sym})^{-1} Y$$

Suppose $F^{(0)} = Y$, then

$$F^{(1)} = \alpha S_{sym} F^{(0)} + (1-\alpha)Y$$
$$= \alpha S_{sym} Y + (1-\alpha)Y$$
$$F^{(2)} = \alpha S_{sym} F^{(1)} + (1-\alpha)Y$$
$$= \alpha^2 S_{sym}^2 Y + (1-\alpha)\alpha S_{sym} Y + (1-\alpha)Y$$
$$F^{(3)} = \alpha S_{sym} F^{(2)} + (1-\alpha)Y$$
$$= \alpha^3 S_{sym}^3 Y + (1-\alpha)\alpha^2 S_{sym}^2 Y + (1-\alpha)\alpha S_{sym} Y + (1-\alpha)Y$$

$$\ldots$$

Thus, by induction,

$$F^{(t)} = \alpha^t S_{sym}^t Y + (1-\alpha)\sum_{i=0}^{t-1}(\alpha S_{sym})^i Y$$

Since for every integer k such that $1 \leq k \leq m$, $D^{(k)-\frac{1}{2}}W^{(k)}D^{(k)-\frac{1}{2}}$ is similar to $D^{(k)-1}W^{(k)}$ which is a stochastic matrix, eigenvalues of $D^{(k)-\frac{1}{2}}W^{(k)}D^{(k)-\frac{1}{2}}$ belong to [-1,1]. Moreover, for every k, $D^{(k)-\frac{1}{2}}W^{(k)}D^{(k)-\frac{1}{2}}$ is symmetric, then $\sum_{k=1}^{m} D^{(k)-\frac{1}{2}}W^{(k)}D^{(k)-\frac{1}{2}}$ is also symmetric. Therefore, by using Weyl's inequality in [10] and the references therein, the largest eigenvalue of $\sum_{k=1}^{m} D^{(k)-\frac{1}{2}}W^{(k)}D^{(k)-\frac{1}{2}}$ is at most the sum of every largest eigenvalues of $D^{(k)-\frac{1}{2}}W^{(k)}D^{(k)-\frac{1}{2}}$ and the smallest eigenvalue of $\sum_{k=1}^{m} D^{(k)-\frac{1}{2}}W^{(k)}D^{(k)-\frac{1}{2}}$ is at least the sum of every smallest eigenvalues of $D^{(k)-\frac{1}{2}}W^{(k)}D^{(k)-\frac{1}{2}}$. Thus, the eigenvalues of $S_{sym} = \frac{1}{m}\sum_{k=1}^{m}D^{(k)-\frac{1}{2}}W^{(k)}D^{(k)-\frac{1}{2}}$ belong to [-1,1]. Moreover, since $0<\alpha<1$, thus

$$\lim_{t \to \infty} \alpha^t S_{sym}^t = 0$$

$$\lim_{t \to \infty} \sum_{i=0}^{t-1}(\alpha S_{sym})^i = (I - \alpha S_{sym})^{-1}$$

Therefore,

$$F^* = \lim_{t \to \infty} F^{(t)} = (1-\alpha)(I - \alpha S_{sym})^{-1}Y$$

Now, from the above formula, we can compute $F^*$ directly.

## 3. Regularization Frameworks

In this section, we will develop the regularization framework for the normalized graph Laplacian based semi-supervised learning iterative version. First, let's consider the error function

$$E(F) = \left\{\frac{1}{2m}\sum_{k=1}^{m}\sum_{i,j=1}^{n}W_{ij}^{(k)}\left\|\frac{F_i}{\sqrt{d_i^{(k)}}} - \frac{F_j}{\sqrt{d_j^{(k)}}}\right\|^2\right\} + \gamma\sum_{i=1}^{n}\|F_i - Y_i\|^2$$

In this error function $E(F)$, $F_i$ and $Y_i$ belong to $R^c$. Please note that c is the total number of protein functional classes, $d_i^{(k)} = \sum_j W_{ij}^{(k)}$, and $\gamma$ is the positive regularization parameter. Hence

$$F = \begin{bmatrix} F_1^T \\ \vdots \\ F_n^T \end{bmatrix} \text{ and } Y = \begin{bmatrix} Y_1^T \\ \vdots \\ Y_n^T \end{bmatrix}$$

Here $E(F)$ stands for the sum of the square loss between the estimated label matrix and the initial label matrix and the smoothness constraint.

Hence we can rewrite $E(F)$ as follows

$$E(F) = trace\big(F^T(I - S_{sym})F\big) + \gamma trace((F - Y)^T(F - Y))$$

Our objective is to minimize this error function. In the other words, we solve

$$\frac{\partial E}{\partial F} = 0$$

This will lead to

$$(I - S_{sym})F + \gamma(F - Y) = 0$$

$$F - S_{sym}F + \gamma F = \gamma Y$$

$$F - \frac{1}{1+\gamma}S_{sym}F = \frac{\gamma}{1+\gamma}Y$$

$$\left(I - \frac{1}{1+\gamma}S_{sym}\right)F = \frac{\gamma}{1+\gamma}Y$$

Let $\alpha = \frac{1}{1+\gamma}$. Hence the solution $F$ of the above equations is

$$F = (1 - \alpha)(I - \alpha S_{sym})^{-1}Y$$

Also, please note that $S_{rw} = \frac{1}{m}\sum_{k=1}^{m} D^{(k)-1} W^{(k)}$ is not the symmetric matrix, thus we cannot develop the regularization framework for the random walk graph Laplacian based semi-supervised learning iterative version.

Next, we will develop the regularization framework for the un-normalized graph Laplacian based semi-supervised learning algorithms. First, let's consider the error function

$$E(F) = \left\{ \frac{1}{2m} \sum_{k=1}^{m} \sum_{i,j=1}^{n} W_{ij}^{(k)} \|F_i - F_j\|^2 \right\} + \gamma \sum_{i=1}^{n} \|F_i - Y_i\|^2$$

In this error function $E(F)$, $F_i$ and $Y_i$ belong to $R^c$. Please note that c is the total number of protein functional classes and $\gamma$ is the positive regularization parameter. Hence

$$F = \begin{bmatrix} F_1^T \\ \vdots \\ F_n^T \end{bmatrix} \text{ and } Y = \begin{bmatrix} Y_1^T \\ \vdots \\ Y_n^T \end{bmatrix}$$

Here $E(F)$ stands for the sum of the square loss between the estimated label matrix and the initial label matrix and the smoothness constraint.

Hence we can rewrite $E(F)$ as follows

$$E(F) = \frac{1}{m} trace\left[ F^T \sum_{k=1}^{m} L^{(k)} F \right] + \gamma trace((F-Y)^T (F-Y))$$

Please note that un-normalized Laplacian matrix of the $k^{th}$ networkis $L^{(k)} = D^{(k)} - W^{(k)}$. Our objective is to minimize this error function. In the other words, we solve

$$\frac{\partial E}{\partial F} = 0$$

This will lead to

$$\frac{1}{m} \sum_{k=1}^{m} L^{(k)} F + \gamma(F-Y) = 0$$

$$\left(\frac{1}{m}\sum_{k=1}^{m} L^{(k)} + \gamma I\right) F = \gamma Y$$

Hence the solution $F$ of the above equations is

$$F = \gamma \left(\frac{1}{m} \sum_{k=1}^{m} L^{(k)} + \gamma I\right)^{-1} Y$$

Similarly, we can also obtain the other form of solution $F$ of the normalized graph Laplacian based semi-supervised learning algorithm as follows (note normalized Laplacian matrix of $k^{th}$ networkis $L_{sym}^{(k)} = I - D^{(k)-\frac{1}{2}} W^{(k)} D^{(k)-\frac{1}{2}}$)

$$F = \gamma \left(\frac{1}{m} \sum_{k=1}^{m} L_{sym}^{(k)} + \gamma I\right)^{-1} Y$$

## 4. Experiments and results
**The Dataset**

The three symmetric normalized, random walk, and un-normalized graph Laplacian based semi-supervised learning are applied to the dataset obtained from [6]. This dataset is composed of 3588 yeast proteins from *Saccharomyces cerevisiae*, annotated with 13 highest-level functional classes from MIPS Comprehensive Yeast Genome Data (Table 1). This dataset contains five networks of pairwise relationships, which are very sparse. These five networks are network created from Pfam domain structure ($W^{(1)}$), co-participation in a protein complex ($W^{(2)}$), protein-protein interaction network ($W^{(3)}$), genetic interaction network ($W^{(4)}$), and network created from cell cycle gene expression measurements ($W^{(5)}$).

The first network, $W^{(1)}$, was obtained from the Pfam domain structure of the given genes. At the time of the curation of the dataset, Pfam contained 4950 domains. For each protein, a binary vector of this length was created. Each element of this vector represents the presence or absence of one Pfam domain. The value of $W_{ij}^{(1)}$ is then the normalization of the dot product between the domain vectors of proteins i and j.

The fifth network, $W^{(5)}$, was obtained from gene expression data collected by [12]. In this network, an edge with weight 1 is created between two proteins if their gene expression profiles are sufficiently similar.

The remaining three networks were created with data from the MIPS Comprehensive Yeast Genome Database (CYGD). $W^{(2)}$ is composed of binary edges indicating whether the given proteins are known to co-participate in a protein complex. The binary edges of $W^{(3)}$ indicate known protein-protein physical interactions. Finally, the binary edges in $W^{(4)}$ indicate known protein-protein genetic interactions.

The protein functional classes these proteins were assigned to are the 13 functional classes defined by CYGD at the time of the curation of this dataset. A brief description of these functional classes is given in the following Table 1.

**Table 1:** 13 CYGD functional classes

|    | Classes |
|----|---------|
| 1  | Metabolism |
| 2  | Energy |
| 3  | Cell cycle and DNA processing |
| 4  | Transcription |
| 5  | Protein synthesis |
| 6  | Protein fate |
| 7  | Cellular transportation and transportation mechanism |
| 8  | Cell rescue, defense and virulence |
| 9  | Interaction with cell environment |
| 10 | Cell fate |
| 11 | Control of cell organization |
| 12 | Transport facilitation |
| 13 | Others |

## Results

In this section, we experiment with the above three methods in terms of classification accuracy performance measure. All experiments were implemented in Matlab 6.5 on virtual machine.

For the comparisons discussed here, the three-fold cross validation is used to compute the accuracy performance measures for each class and each method. The accuracy performance measure Q is given as follows

$$Q = \frac{True\ Positive + True\ Negative}{True\ Positive + True\ Negative + False\ Positive + False\ Negative}$$

True Positive (TP), True Negative (TN), False Positive (FP), and False Negative (FN) are defined in the following table 2

**Table 2:** Definitions of TP, TN, FP, and FN

|  |  | Predicted Label |  |
|---|---|---|---|
|  |  | Positive | Negative |
| Known Label | Positive | True Positive (TP) | False Negative (FN) |
|  | Negative | False Positive (FP) | True Negative (TN) |

In these experiments, parameter $\alpha$ is set to 0.85 and $\gamma = 1$. For this dataset, the third table shows the accuracy performance measures of the three methods applying to integrated network for 13 functional classes

**Table 3:** Comparisons of symmetric normalized, random walk, and un-normalized graph Laplacian based methods using integrated network

| Functional Classes | Accuracy Performance Measures (%) Integrated Network | | |
|---|---|---|---|
|  | Normalized | Random Walk | Un-normalized |
| 1 | 76.87 | 76.98 | **77.20** |
| 2 | **85.90** | 85.87 | 85.81 |
| 3 | **78.48** | **78.48** | 77.56 |
| 4 | **78.57** | 78.54 | 77.62 |
| 5 | 86.01 | 85.95 | **86.12** |
| 6 | 80.43 | **80.49** | 80.32 |
| 7 | **82.02** | 81.97 | 81.83 |
| 8 | **84.17** | 84.14 | **84.17** |
| 9 | 86.85 | 86.85 | **86.87** |
| 10 | **80.88** | 80.85 | 80.52 |
| 11 | 85.03 | 85.03 | **85.92** |
| 12 | 87.49 | 87.46 | **87.54** |
| 13 | **88.32** | **88.32** | **88.32** |

From the above table 3, we recognized that the symmetric normalized and un-normalized graph Laplacian based semi-supervised learning methods slightly perform better than the random walk graph Laplacian based semi-supervised learning method.

Next, we will show the accuracy performance measures of the three methods for each individual network $W^{(k)}$ in the following tables:

**Table 4:** Comparisons of symmetric normalized, random walk, and un-normalized graph Laplacian based methods using network $W^{(1)}$

| Functional Classes | Accuracy Performance Measures (%) Network $W^{(1)}$ | | |
|---|---|---|---|
| | Normalized | Random Walk | Un-normalized |
| 1 | 64.24 | 63.96 | **64.30** |
| 2 | 71.01 | 71.07 | **71.13** |
| 3 | 63.88 | 63.66 | **63.91** |
| 4 | **65.55** | 65.41 | 65.47 |
| 5 | 71.35 | **71.46** | 71.24 |
| 6 | 66.95 | 66.69 | **67.11** |
| 7 | **67.89** | 67.70 | 67.84 |
| 8 | 69.29 | 69.29 | **69.31** |
| 9 | 71.49 | 71.40 | **71.52** |
| 10 | 65.30 | 65.47 | **65.50** |
| 11 | 70.09 | 70.04 | **70.12** |
| 12 | **72.71** | 72.66 | 72.63 |
| 13 | **72.85** | 72.77 | **72.85** |

**Table 5:** Comparisons of symmetric normalized, random walk, and un-normalized graph Laplacian based methods using network $W^{(2)}$

| Functional Classes | Accuracy Performance Measures (%) Network $W^{(2)}$ | | |
|---|---|---|---|
| | Normalized | Random Walk | Un-normalized |
| 1 | **24.64** | **24.64** | **24.64** |
| 2 | **27.84** | **27.84** | 27.79 |
| 3 | **23.16** | **23.16** | 23.08 |

21

| | | | |
|---|---|---|---|
| 4 | **22.60** | **22.60** | 22.52 |
| 5 | **26.37** | **26.37** | 26.23 |
| 6 | **24.39** | **24.39** | 24.19 |
| 7 | 26.11 | 26.11 | **26.37** |
| 8 | **27.65** | **27.65** | 27.62 |
| 9 | **28.43** | **28.43** | 28.34 |
| 10 | **25.81** | **25.81** | 25.22 |
| 11 | **27.01** | **27.01** | 25.98 |
| 12 | **28.43** | **28.43** | 28.40 |
| 13 | **28.54** | **28.54** | **28.54** |

**Table 6:** Comparisons of symmetric normalized, random walk, and un-normalized graph Laplacian based methods using network $W^{(3)}$

| Functional Classes | Accuracy Performance Measures (%) Network $W^{(3)}$ | | |
|---|---|---|---|
| | Normalized | Random Walk | Un-normalized |
| 1 | **29.63** | 29.57 | 29.40 |
| 2 | **34.11** | **34.11** | 33.95 |
| 3 | **27.93** | 27.90 | 27.70 |
| 4 | 28.51 | 28.48 | **28.57** |
| 5 | **34.03** | **34.03** | 33.92 |
| 6 | **30.57** | 30.55 | 30.04 |
| 7 | **32.08** | **32.08** | 32.02 |
| 8 | **33.05** | 33.03 | 32.92 |
| 9 | **33.78** | **33.78** | 33.75 |
| 10 | **30.18** | **30.18** | 29.99 |
| 11 | **32.64** | **32.64** | 32.53 |
| 12 | **34.53** | **34.53** | 34.45 |
| 13 | **34.48** | **34.48** | 34.31 |

**Table 7:** Comparisons of symmetric normalized, random walk, and un-normalized graph Laplacian based methods using network $W^{(4)}$

| Functional Classes | Accuracy Performance Measures (%) Network $W^{(4)}$ | | |
|---|---|---|---|
| | Normalized | Random Walk | Un-normalized |
| 1 | **18.31** | 18.28 | 18.26 |
| 2 | **20.93** | 20.90 | 20.88 |
| 3 | **18.09** | 18.06 | **18.09** |
| 4 | **18.39** | **18.39** | **18.39** |
| 5 | **21.07** | **21.07** | 21.04 |
| 6 | **18.98** | **18.98** | 18.90 |
| 7 | **18.73** | **18.73** | 18.67 |
| 8 | **19.90** | **19.90** | 19.62 |
| 9 | **20.04** | **20.04** | 19.93 |
| 10 | **17.31** | 17.28 | 17.17 |
| 11 | **19.18** | **19.18** | 19.09 |
| 12 | 20.54 | 20.54 | **20.57** |
| 13 | **20.54** | **20.54** | 20.48 |

**Table 8:** Comparisons of symmetric normalized, random walk, and un-normalized graph Laplacian based methods using network $W^{(5)}$

| Functional Classes | Accuracy Performance Measures (%) Network $W^{(5)}$ | | |
|---|---|---|---|
| | Normalized | Random Walk | Un-normalized |
| 1 | 26.45 | 26.45 | **26.51** |
| 2 | **29.21** | **29.21** | **29.21** |

| 3 | 25.89 | 25.78 | **25.92** |
| 4 | **26.76** | 26.62 | **26.76** |
| 5 | **29.18** | **29.18** | **29.18** |
| 6 | **27.42** | 27.23 | **27.42** |
| 7 | **28.21** | 28.18 | 28.01 |
| 8 | 28.51 | **28.54** | **28.54** |
| 9 | **29.71** | 29.68 | 29.65 |
| 10 | 26.81 | 26.95 | **27.01** |
| 11 | 28.79 | 28.82 | **28.85** |
| 12 | **30.16** | 30.13 | **30.16** |
| 13 | **30.18** | 30.16 | **30.18** |

From the above tables, we easily see that the un-normalized (i.e. the published) and normalized graph Laplacian based semi-supervised learning methods slightly perform better than the random walk graph Laplacian based semi-supervised learning method using network $W^{(1)}$ and $W^{(5)}$. For $W^{(2)}$, $W^{(3)}$, and $W^{(4)}$, the random walk and the normalized graph Laplacian based semi-supervised learning methods slightly perform better than the un-normalized (i.e. the published) graph Laplacian based semi-supervised learning method. $W^{(2)}$, $W^{(3)}$, and $W^{(4)}$ are all three networks created with data from the MIPS Comprehensive Yeast Genome Database (CYGD).

Moreover, the accuracy performance measures of all three methods for $W^{(2)}$, $W^{(3)}$, $W^{(4)}$, and $W^{(5)}$ are un-acceptable since they are worse than random guess. Again, this fact occurs due to the sparseness of these four networks.

For integrated network and every individual network except $W^{(1)}$, we recognize that the symmetric normalized graph Laplacian based semi-supervised learning method performs slightly better than the other two graph Laplacian based methods.

Finally, the accuracy performance measures of these three methods for the integrated network are much better than the best accuracy performance measure of these three methods for individual network. Due to the sparseness of the networks, the accuracy performance measures for individual networks W2, W3, W4, and W5 are unacceptable. They are worse than random guess. The best accuracy performance measure of these three methods for individual network will be shown in the following supplemental table.

**Supplement Table:** Comparisons of un-normalized graph Laplacian based methods using network $W^{(1)}$ and integrated network

| Functional Classes | Accuracy Performance Measures (%) | |
|---|---|---|
| | Integrated network (un-normalized) | Best individual network $W^{(1)}$ (un-normalized) |
| 1 | 77.20 | 64.30 |
| 2 | 85.81 | 71.13 |
| 3 | 77.56 | 63.91 |
| 4 | 77.62 | 65.47 |
| 5 | 86.12 | 71.24 |
| 6 | 80.32 | 67.11 |
| 7 | 81.83 | 67.84 |
| 8 | 84.17 | 69.31 |
| 9 | 86.87 | 71.52 |
| 10 | 80.52 | 65.50 |
| 11 | 84.92 | 70.12 |
| 12 | 87.54 | 72.63 |
| 13 | 88.32 | 72.85 |

## 5. Conclusion

The detailed iterative algorithms and regularization frameworks for the three normalized, random walk, and un-normalized graph Laplacian based semi-supervised learning methods applying to the integrated network from multiple networks have been developed. These three methodsare successfully applied to the protein function prediction problem (i.e. classification problem). Moreover, the comparison of the accuracy performance measures for these three methods has been done.

These three methods can also be applied to cancer classification problems using gene expression data.

Moreover, these three methods can not only be used in classification problem but also in ranking problem. In specific, given a set of genes (i.e. the queries) making up a protein complex/pathways or given a set of genes (i.e. the queries) involved in a specific disease (for e.g. leukemia), these three methods can also be used to find more potential members of the complex/pathway or more genes involved in the same disease by ranking genes in gene co-expression network (derived from gene expression data) or the protein-protein interaction network or the integrated network of them. The genes with the highest rank then will be selected and then checked by biologist experts to see if the extended genes in fact belong to the same complex/pathway or are involved in the same disease. These problems are also called complex/pathway membership determination and biomarker discovery in cancer classification. In cancer classification problem, only the sub-matrix of the gene expression data of the extended gene list will be used in cancer classification instead of the whole gene expression data.

Finally, to the best of my knowledge, the normalized, random walk, and un-normalized hypergraph Laplacian based semi-supervised learning methods have not been applied to the protein function prediction problem. These methods applied to protein function prediction are worth investigated since [11] have shown that these hypergraph Laplacian based semi-supervisedlearning methods outperform the graph Laplacian based semi-supervised learning methods in text-categorization and letter recognition.